\documentclass[conference]{IEEEtran}
\IEEEoverridecommandlockouts
\usepackage{cite}
\usepackage{amsmath,amssymb,amsfonts}
\usepackage{algorithmic}
\usepackage{graphicx}
\usepackage{textcomp}
\usepackage{xcolor}
\usepackage{multirow}
\usepackage{array}
\usepackage{times}
\usepackage{latexsym}
\usepackage{enumitem}
\usepackage{booktabs}
\usepackage{hyperref}

\def\BibTeX{{\rm B\kern-.05em{\sc i\kern-.025em b}\kern-.08em
    T\kern-.1667em\lower.7ex\hbox{E}\kern-.125emX}}
\begin{document}

\title{Dialogue Inspectional Summarization with \\ Factual Inconsistency Awareness}
\author{
\IEEEauthorblockN{
    \textbf{Leilei Gan\textsuperscript{\rm 1,\textsection},Yating Zhang\textsuperscript{\rm 1,\textsection}, Kun Kuang\textsuperscript{\rm 1,\textsection}, Lin Yuan\textsuperscript{\rm 1,\textsection}} \\
    \textbf{Shuo Li\textsuperscript{\rm 3}, Changlong Sun\textsuperscript{\rm 2}, Xiaozhong Liu\textsuperscript{\rm 4}, Fei Wu\textsuperscript{\rm 1,\textdagger}}
}
\IEEEauthorblockA{\textit{}
\textsuperscript{1}Zhejiang University, Hangzhou, China, \textsuperscript{2}Alibaba Group, Hangzhou, China 
\\
\textsuperscript{3}National University of Singapore, Singapore, Singapore, \textsuperscript{4}Indiana University, Indiana, USA\\
{\tt leileigan@zju.edu.cn, yatingz89@gmail.com, changlong.scl@taobao.com}\\
{\tt kunkuang@zju.edu.cn, liu237@indiana.edu, wufei@zju.edu.cn}
}}

\maketitle
\begingroup\renewcommand\thefootnote{\textsection}
\footnotetext{Equal contribution}
\endgroup

\begingroup\renewcommand\thefootnote{\textdagger}
\footnotetext{Corresponding Author}
\endgroup

\begin{abstract}
    Dialogue summarization has been extensively studied and applied, where the prior works mainly focused on exploring superior model structures to align the input dialogue and the output summary. 
    However, for professional dialogues (e.g., legal debate and medical diagnosis), semantic/statistical alignment can hardly fill the logical/factual gap between input dialogue discourse and summary output with external knowledge. 
    In this paper, we mainly investigate the factual inconsistency problem for Dialogue Inspectional Summarization (DIS) under non-pretraining and pretraining settings. 
    An innovative end-to-end dialogue summary generation framework is proposed with two auxiliary tasks: Expectant Factual Aspect Regularization (EFAR) and Missing Factual Entity Discrimination (MFED). 
    Comprehensive experiments demonstrate that the proposed model can generate a more readable summary with accurate coverage of factual aspects as well as informing the user with potential missing facts detected from the input dialogue for further human intervention.
\end{abstract}


\section{Introduction}
\label{Introduction}
Summarization-based algorithms have enabled a broad spectrum of applications, such as auto-abbreviated news and retrieval outcomes~\cite{gerani2014abstractive, grusky2018newsroom} to assist users in consuming lengthy document effectively. 
Generally, summarization methods can be divided into extractive summarization and abstractive summarization. 
The former selects sentences from the original document to compose the summaries~\cite{mihalcea2004textrank,DBLP:conf/aaai/NallapatiZZ17}, while the latter generates the summary texts one word by word based on the understanding of the documents~\cite{jing-mckeown-2000-cut,rush-etal-2015-neural,see2017get}.
In recent years, with the renaissance of neural networks, deep learning based methods have achieved the state-of-the-art results on standard benchmarks for both abstractive and extractive summarization~\cite{rush2015a,gehrmann2018bottom, hsu-etal-2018-unified, celikyilmaz-etal-2018-deep}.

Recently, dialogue summarization~\cite{goo2018abstractive,liu2019automatic,feng2021survey}  also raised much attention of the research community, thanks to the development of automatic speech recognition techniques. 
Different from summarizing the plain documents, the goal of dialogue summarization is to generate summaries for dialogues. 
Exemplar applications can be found in meetings, customer services and judicial trials. 
However, there exists two special challenges for dialogue summarization. The first one is that multi-role dialogue is more complicated due to the interactions among the different parties. Enhanced representation of the atomic components (e.g., utterance and role) of the dialogue prequalifies summary generation optimization. 

\begin{figure}[t]
\centering
\includegraphics[width=\columnwidth]{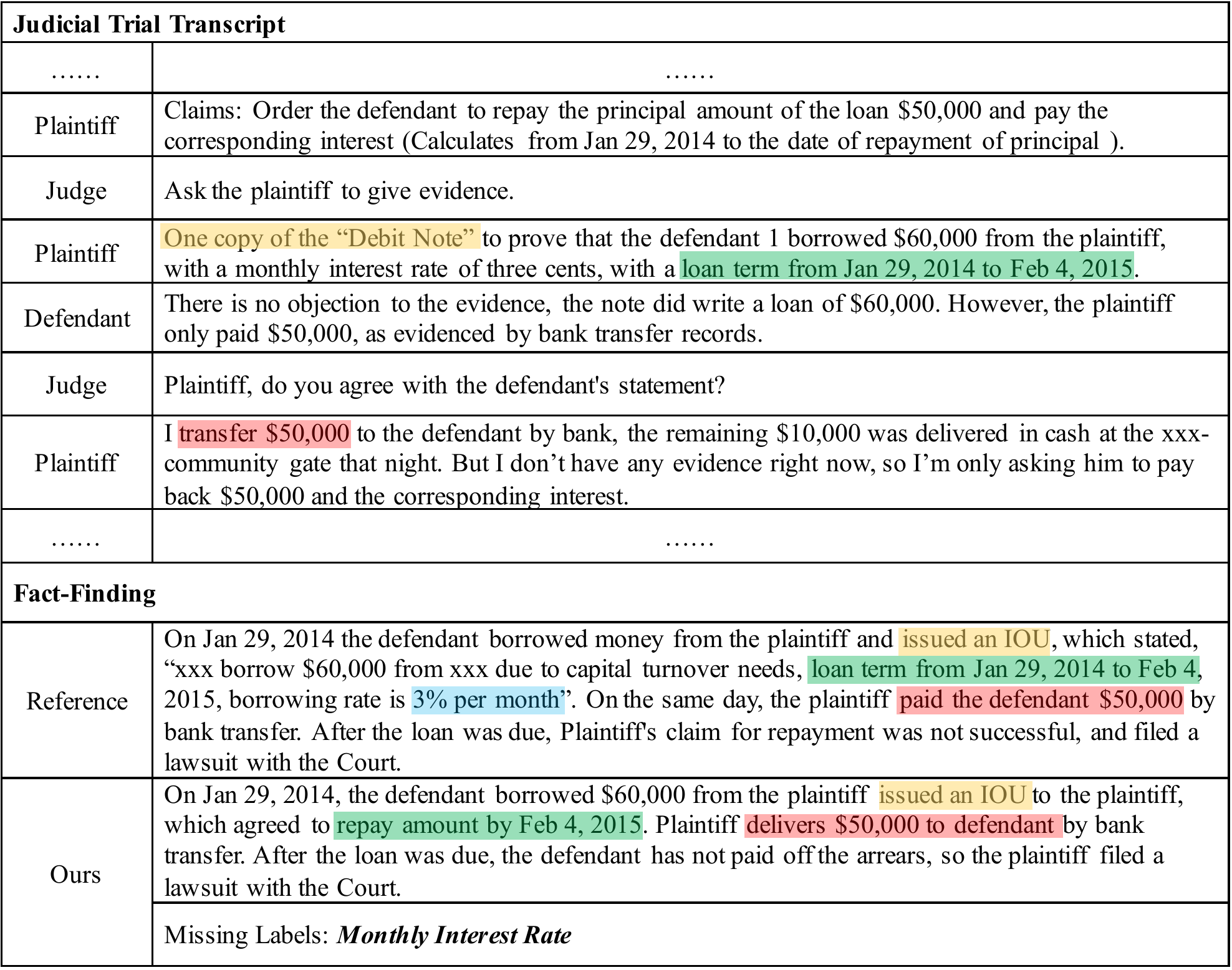}
\caption{An example of judicial trial transcript, as well as summaries written by judge and generated by our method. Different colors represent different factual aspects. Our result informs the users with missing entity (Monthly Interest Rate) for further human intervention.}
\label{IntroCase}
\end{figure}

Second, despite the nature of multi-role dialogues that are noisy and tedious for many circumstances (e.g., a 1-2 hour judicial trial can host 10-30 thousand words), sometimes, there exists a logical/factual gap between the input dialogue and the expected output summary. 
For instance, in the court debate scenario~\cite{DBLP:conf/cikm/DuanZYZLWWZS019}, the judge summarizes the case narrative not only based on facts recognized from the court debate during the trial but also relying on the evidence or materials submitted by the litigants. 
Similarly, in the medical inquiry process between the doctor and patient, the doctor concludes the medical record based on their conversation on symptoms along with a series of medial reports (e.g., blood test, X-Ray and CT results). In such context, the current summarization framework \cite {see2017get, paulus2018a} by enforcing alignment between the input (with dialogue discourse) and the output (with extra knowledge from external resource) can result in the mismatch and misalignment in terms of the critical factual aspects and logics in the generated summary, i.e., semantic/statistical alignment can hardly fill the logical/factual gaps in dialogue summarization. 

Motivated by such observation, in this paper, we particularly investigate the factual inconsistency problem and propose a novel framework: \textbf{Dialogue Inspectional Summarization (DIS)} under both non-pretraining and pretraining settings. 
Specifically, in the non-pretraining setting, we design a hierarchical dialogue encoder involving role information to accommodate long context and multiple turns among the multiple roles. 
Rather than directly aligning the input dialogue and its summary within the generation framework, we additionally propose two auxiliary tasks in the way of joint learning: \textbf{Expectant Factual Aspect Regularization (EFAR)} can estimate the factual aspects to be contained in the summary so as to make the model emphasize on the factual coverage of logic reasoning; and \textbf{Missing Factual Entity Discrimination (MFED)} predicts the missing aspects which discover/alarm the factual gap between the input and the output. 
DIS allows the users (e.g., judge and doctor) to further improve the summary by referring to other materials (e.g., evidence) according to the detected missing aspect-based entities. 
In the pretraining setting, we further investigate the factual inconsistency problem of pre-trained summarization models by equipping DIS with a pre-trained model PEGASUS~\cite{DBLP:conf/icml/ZhangZSL20}.

Figure~\ref{IntroCase}, for instance, shows a judicial trial transcript and the summary generated by our model. The predicted aspects and the missing aspects are produced by the two auxiliary tasks respectively. Such complementary results can help the professional users better understand the generated content in terms of the factual aspects coverage and the possible omissions that are not mentioned in the input dialogue but essential for writing a comprehensive summary.

To sum up, our contributions are as follows:
\begin{itemize}
    \item To the best of our knowledge, this work is the first attempt for dialogue inspectional summarization (DIS) by addressing the factual inconsistency between the input dialogue and the expected output summary. We particularly conduct experiments and validate our hypothesis on the judicial trial scenario.
    \item We present an end-to-end DIS framework in both non-pretraining and pretraining settings, which is supervised by two auxiliary tasks: EFAR and MFED. Comprehensive experiments demonstrate that the proposed model can generate a more readable summary with high coverage of factual aspects as well as informing users with potential missing facts detected from the input dialogue.
    \item We benchmark the DIS dataset in the judicial domain, where the input is the real civil trial debate data. The summaries come from the factfinding paragraphs extracted from the judgment documents of the corresponding cases. The factual aspect annotation is conducted on the output summary by five legal experts. To motivate other scholars to investigate this novel and essential problem, we will make the experiment dataset publicly available (while removing the sensitive information).
\end{itemize}





\section{Problem Formulation}
Dialogue inspectional summarization is the task of condensing the factual information mentioned in the dialogue to a shorter version summary, along with two auxiliary tasks: Expectant Factual Aspect Regularization (EFAR) and Missing Factual Entity Discrimination (MFED).

Formally, each case contains a multi-role dialogue $D$ and an inspectional summary $F$. 
The dialogue $D = (u_1, u_2, ..., u_L)$ includes $L$ utterances where each utterance $u_i$ is composed of a sequence of $l$ words $u_i = (w_{i1}, w_{i2}, ..., w_{il})$ and its speaker role $r_i$. 
The summary of the dialogue is $F=(f_1, f_2, ..., f_N)$, where $N$ is the number of tokens.

During the annotation process, we define a set of factual aspects $A = (a_1, a_2, ..., a_K)$ and a set of factual entities $E = (e_1, e_2, ..., e_M)$ respectively, with the assistance of domain experts. 
For each sample, we annotate the expectant value $\hat{y_{i}^{a}} \in \{0, 1\}$ for each aspect $a_i$ in set $A$ and missing value $\hat{y_{i}^{m}} \in \{0, 1\}$ for each entity $e_i$ in set $E$. 
To be specific, aspect expectant value indicates whether the aspect should be considered in the case. 
Entity missing value indicates whether the information is related to the case but was omitted in the dialogue, thus leading to factual inconsistency between dialogue and inspectional summary.

In more detail, the defined 12 factual aspects are Limitation of Action, Notice of Repayment, Written Loan Agreement, Nature of Loan, 
Guarantee Liability, Agreed Rate/Interest, Repayment Period, Loan Period, Breach Clause, Repayment Behavior, Delivery, Debt.
And the defined 14 expectant aspects are Loan Amount, Loan Period, Loan Start Date, Loan End Date, Repayment Date, Repayment of Principal, Repayment of Interest, Penalty/Overdue Interest, Outstanding Principal,
Delivery Date, Delivery Amount, Annual Interest Rate, Monthly Interest Rate, Overdue Interest Rate.
More details can refer to the released two complete samples of our dataset\footnote{\url{https://github.com/anonymous-tmp/anonymous-1}}.
\section{Method}
\begin{figure*}[t]
\centering
\includegraphics[scale=0.8]{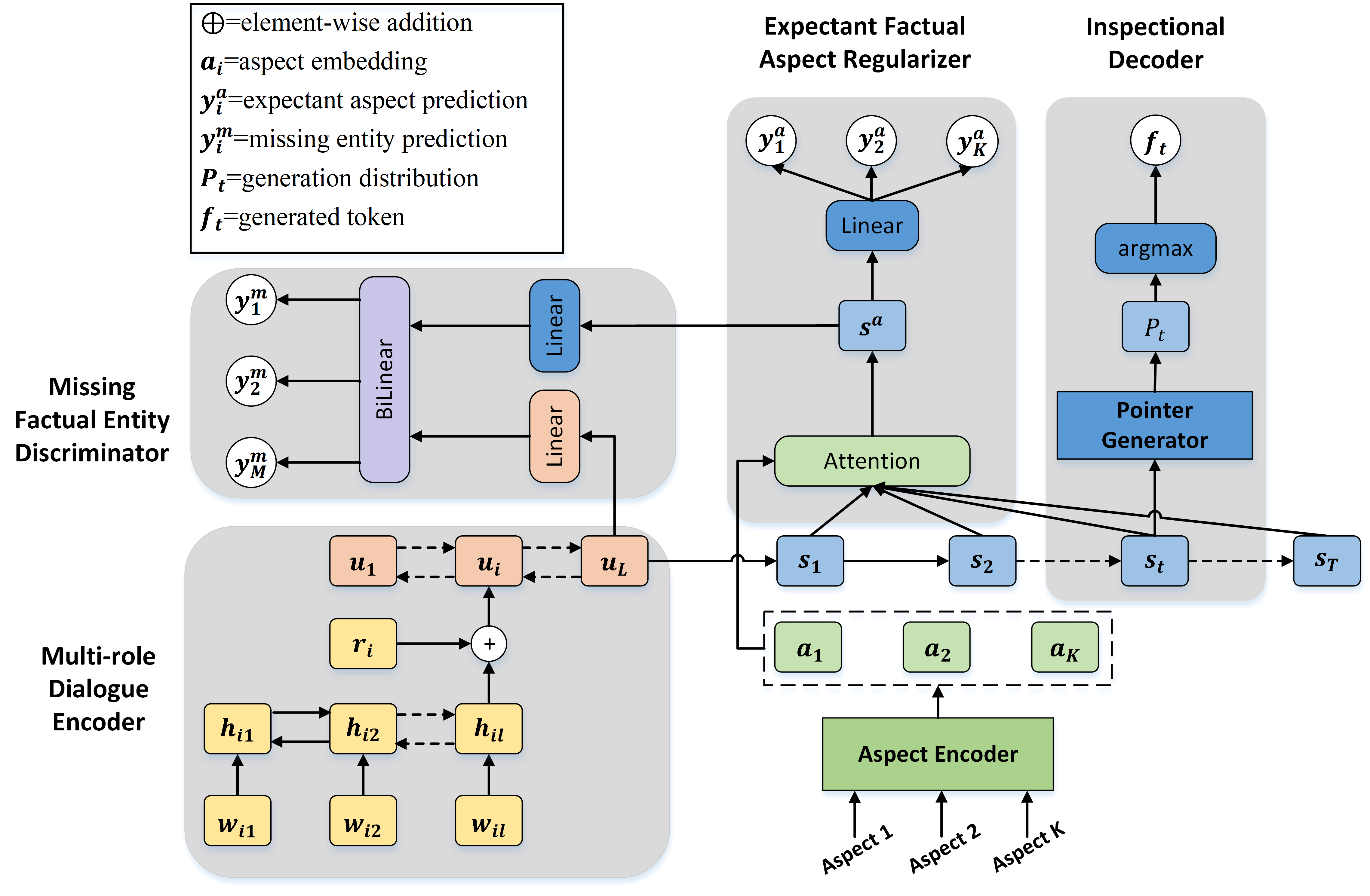}
\caption{DIS framework overview. The framework is modularized as Multi-role Dialogue Encoder, Inspectional Decoder, Expectant Factual Aspect Regularizer, and Missing Factual Entity Discriminator.}
\label{modelPic}
\end{figure*}
In this section, we introduce our dialogue summary generation framework, Dialogue Inspectional Summarization (DIS). 
DIS exploits domain knowledge to relieve the factual inconsistency problem and improve the credibility of the generated summary. 
We first describe it in a non-pretraining setting, which consists of Dialogue Encoder and Inspectional Decoder.
Then we introduce the encoder-decoder model in a pre-trained setting.
Third, we introduce how to combine the non-pretraining and pretraining summarization model with two auxiliary tasks: (1) Expectant Factual Aspect Regularizer, and (2) Missing Factual Entity Discriminator.
Finally, the objective function is described for parameters Optimization.

\subsection{Dialogue Encoder}
Considering the contextual structure of dialogues, we employ hierarchical encoders with role information integrated for dialogue representative learning. We first encode each utterance and then encode the entire dialogue. At utterance level, we use utterance encoder, notated as ${\rm \bf Enc^{U}}$, to represent the local semantics of each utterance:
\begin{equation}
    {\bf h}_{ij} = {\rm \bf Enc^{U}}({\bf w}_{ij}), \quad j \in \left[ 1, l \right]
\end{equation}
where ${\bf h}_{ij}$ are encoder hidden states corresponding to the token $j$ in utterance $i$, ${\bf w}_{ij}$ is the dense word embedding, and $l$ is the maximum length of utterance.

In general, utterances of different characters contain different angles of factual information in a multi-role dialogue scenario. For example, in a trial debate, the plaintiffs tend to state the facts while the defendants deny them. Meanwhile, the presiding judge asks questions about the factual details to advance the trial. Inspired by their different functions in the dialogue, we integrate the role information into utterance representation. To be specific, we apply the role embedding layer to learn the representation ${\bf r}$ for each type of the role (presiding judge, plaintiff, and defendant), corresponded to each utterance:
\begin{equation}
    {\bf u}_{i}^{\bf local} = {\bf h}_{i} + {\bf r}_i 
\end{equation}
where ${\bf u}_{i}^{\bf local}$ represents the local semantics of utterance under the specific speaker role. The utterance hidden state ${\bf h}_i$ is the last state of ${\bf h}_{ij}$, where $j \in \left[ 1, l \right]$.

We then employ dialogue encoder, notated as ${\rm \bf Enc^{D}}$, to learn the contextual representations ${\bf u}_i$ of each utterance from the dialogue level:
\begin{equation}
    {\bf u}_{i} = {\rm \bf Enc^{D}}({\bf u}_{i}^{\bf local}), \quad i \in \left[ 1, L \right]
\end{equation}
where $L$ is the maximum utterance number of dialogue. Dialogue embedding ${\bf d}$ is the last state of ${\bf u}_i$, where $i \in [1, L]$.

In our implementation, both ${\rm \bf Enc^{D}}$ and ${\rm \bf Enc^{U}}$ are single layer bidirectional LSTM\footnote{We have also tested attentional LSTM and Transformer as the encoder, but they did not give better performance.}. 
We concatenate the forward and backward LSTM states to represent encoder hidden states:
\begin{equation}
    {\bf h} = \left[ {\overrightarrow{\bf h}}, {\overleftarrow{\bf h}} \right]
\end{equation}

\subsection{Inspectional Decoder}
We propose an Inspectional Decoder for generating summaries. Inspectional Decoder generates the summary via pointing mechanism, while the Expectant Factual Aspect Regularizer tries to ensure the factual consistency from aspect level.

From the perspective of bionics, humans tend to write a draft before focusing on factual aspects. We treat the Inspectional Decoder as a drafter, whose states need further regularized by the aspect-aware module. At each decoding timestep $t$, context vector ${\bf c_{t}}$ is computed to represent the attended information from dialogue:
\begin{equation}
    {\bf c}_{t} = \sum_{i=1}^{L}\sum_{j=1}^{l} \alpha_{ij}^{t} {\bf h}_{ij}
\end{equation}
where $\alpha_{ij}^{t}$ shows the corresponding attention distribution on encoder state ${\bf h}_{ij}$. The attended probability $\alpha_{ij}^{t}$ is calculated given the decoder state ${\bf s}_t$:
\begin{equation}
\begin{aligned}
    &\alpha_{ij}^{t} = \mathop{\rm softmax}\limits_{(i,j)} ({\rm score}({\bf s}_t, {\bf h}_{ij})) \\
    &{\rm score}({\bf s}_t, {\bf h}_{ij}) = ({\bf s}_t)^T{\bf h}_{ij}
\end{aligned}
\end{equation}

The context vector, together with the decoder state, are concatenated to produce the generated probability distribution over the fixed size vocabulary:
\begin{equation}
    P_v = \mathop{\rm softmax}(W_v([{\bf s}_t, {\bf c}_t]) + b_v)
\end{equation}
where $W_v$ and $b_v$ are learnable parameters. Following \cite{see2017get}, we employ the pointing mechanism in the decoder to directly copy tokens from the dialogue, along with generating tokens from a fixed vocabulary. Generation probability $p_g$ is computed as:
\begin{equation}
    {p_g} = \sigma(W_c{\bf c}_t + W_s{\bf s}_t + W_f{\bf f}_t + b_{g})
\end{equation}
where $W_c$, $W_s$, $W_f$, $b_{g}$ are learnable parameters and ${\bf f}_t$ is the decoder input from reference summary $\{ f_1, f_2, ..., f_N\}$. $N$ is the maximum decoding length. With the pointing mechanism, we calculate the probability of token $w$ as follows:
\begin{equation}
    P_t(w)= {p_g} {P_v}(w) + (1-{p_g}) \sum_{(i,j): w_{ij}=w} \alpha_{ij}
\end{equation}

With the pointing mechanism integrated, the decoder can directly copy tokens from dialogue, making generated summary more accurate and relevant in factual details.

\subsection{Pre-trained Encoder-Decoder Model for Dialogue Inspectional Summarization}
In this section, we further investigate whether the proposed dialogue inspectional summarization can benefit the pre-trained encoder-decoder model.

Specifically, we take PEGASUS \cite{DBLP:conf/icml/ZhangZSL20} as the encoder-decoder model. 
PEGASUS is the state-of-the-art abstractive summarization model based on Transformer, which is pre-trained on a large-scale corpora by generating the selected important sentences as self-supervised tasks.
We directly replace the multi-role dialogue encoder with PEGASUS Transformer-based encoder, and the inspectional decoder with the PEGASUS decoder.

Formally, given the input dialogue $D = (u_1, u_2, ..., u_L)$, the PEGASUS encoder concatenates the utterances in the dialogue and encodes the dialogue into deep contextual hidden states as:
\begin{equation}
   {\bf h} = \textbf{PEGASUS}_\text{Enc}(D)
\end{equation}

We generate the decode state ${\bf s_t}$ based on the dialogue hidden states ${\bf h}$ and the previous generated summary text $\hat{F}$ as follows:
\begin{equation}
    {\bf s_{t}} = \textbf{PEGASUS}_\text{Dec}({\bf h}, \hat{F}_{1:t-1})
\end{equation}

Based on the multi-role dialogue encoder-decoder model and pre-trained encoder-decoder models, we next introduce how to relieve the factual inconsistency problem with two auxiliary tasks.

\subsection{Expectant Factual Aspect Regularizer}
\label{aspect_section}
When writing formal documents like the legal verdict, people always carefully review their drafts to ensure that there are no inconsistencies in expectant aspects. Inspired by the process, we propose Expectant Factual Aspect Regularizer to verify the consistency on the aspect level.

For each aspect ${\boldsymbol{a}_{i}}$, we use aspect encoder to obtain its semantic embedding ${\bf a}_i$. 
The encoder ${\rm \bf Enc^{A}}$ is a single layer bidirectional LSTM to represent the aspect description text:
\begin{equation}
    {\boldsymbol{a}_{i}} = {\rm \bf Enc^{A}}(a_i)
\end{equation}

We then produce a weighted sum of the decoder hidden states, known as the aspect-aware decoder state ${\bf s^{\boldsymbol{a}}}$: 
\begin{equation}
\begin{aligned}
    &{\bf s^{\boldsymbol{a}}} = \frac{1}{K}\sum_{i=1}^{K}\sum_{t=1}^{T}\alpha_{it}^{asp}{\bf s}_t \\
    &\alpha_{it}^{asp} = \mathop{\rm softmax}\limits_{t}({\rm score}({\boldsymbol{a}_{i}}, {\bf s}_{t}))
\end{aligned}
\end{equation}
where $K$ is the number of factual aspects and ${\rm score}$ function uses additive attention:
\begin{equation}
    {\rm score}({\boldsymbol{a}_{i}}, {\bf s}_{t}) = v^T{\rm tanh}({\rm linear}({\boldsymbol{a}_{i}}, {\bf s}_t))
\end{equation}

Finally, we feed ${\bf s^{\boldsymbol{a}}}$ into a 3-layer classifier to predict the expectant aspects:
\begin{equation}
    {\bf y^{\boldsymbol{a}}} = \sigma(\mathcal{F}^{a}({\bf{s^{\boldsymbol{a}}}}))
\end{equation}
where $\mathcal{F}^{a}$ is the notation of linear layers and ${\bf y^{\boldsymbol{a}}} \in \mathbb{R}^{K \times 1}$ indicates the related probability of $K$ aspects.

\subsection{Missing Factual Entity Discriminator}
\label{missing_section}
As mentioned in section~\ref{Introduction}, there always exists factual inconsistencies between the dialogue and reference summary. In the Seq2Seq framework, inconsistencies mislead the decoder to generate incorrect factual details. Missing Factual Entity Discriminator tries to detect the inconsistencies, thus mitigating the problem.

Motivated by the observation, we design the discriminator to classify whether the factual entity $e_i$ is missing in the conversation. In real applications, human summarizers can refer the predictions to complete generated text based on additional information.

Intuitively, we view inconsistency as the factual divergence between source content and target content, so we use bilinear layer as the classifier. $\mathcal{F}^{d}$ and $\mathcal{F}^{s}$ represent transformation by linear layers:
\begin{equation}
    {\bf y^m} = \sigma(\mathcal{F}^{d}({\bf d})W_m\mathcal{F}^{s}({\bf s^{\boldsymbol{a}}})+ b_m)
\end{equation}
where ${\bf y^m} \in \mathbb{R}^{M \times 1}$ indicates the missing probability of $M$ entities, $W_m$ and $b_m$ are trainable parameters.

\subsection{Parameter Optimization}
As for the summary generation task, we use negative log-likelihood error computed over all the timesteps:
\begin{equation}
    \mathcal{L}_{g} = \frac{1}{T}\sum_{t=1}^{T}-\log{P(f_t)}
\end{equation}
where $P(f_t)$ is the generation probability of the target token $f_t$.

The two auxiliary tasks introduced in section~\ref{aspect_section} and ~\ref{missing_section} are both multi-label binary classification task, so we use binary cross entropy loss when computing $\mathcal{L}_{c}(y^a, \hat{y^a})$ and $\mathcal{L}_{c}(y^m, \hat{y^m})$:
\begin{equation}
    \mathcal{L}_{c}(y, \hat{y}) = \frac{1}{N}\sum_{i=1}^{N}-y_i\ln{\hat{y_i}} - (1-y_i)\ln{(1-\hat{y_i})}
\end{equation}

Finally, we combine the generation loss and classification loss as the final loss $\mathcal{L}$ to optimize the parameters:
\begin{equation}
    \mathcal{L} = \mathcal{L}_{g} + \lambda_a \mathcal{L}_{c}(y^a, \hat{y^a}) + \lambda_m \mathcal{L}_{c}(y^m, \hat{y^m})
\end{equation}
where $\lambda_a$ and $\lambda_m$ are hyperparameters to balance the main task and auxiliary tasks. 

We name the non-pretraining model as {\bf DIS} and the pre-trained based DIS as {\bf PEGASUS-DIS}.
\section{Experimental Settings}

\subsection{Dataset}
For the experiment, we collected more than 300,000 court trial records of civil Private Loan Disputes (PLD) cases, paired with their corresponding verdicts. 
Legal experts helped us introduce 12 factual aspects and 14 types of factual entity, which are discriminative in PLD cases, and labeled 45,531 cases, i.e., for each case, we annotate the correlation of each aspect and the consistency of each entity. 
The aspect annotation indicates whether it will be considered in the court investigation. 
The entity annotation indicates whether its information is consistent between the trial and the verdict. 
In the inter-rater reliability measurement, the annotators achieved the Kappa coefficient as 0.9 (substantial agreement) after training.

We constructed the dataset by extracting the factfinding part from verdicts and filtered out those cases with a kind of fixed mode factfinding. 
The statistics of the dataset are shown in Table~\ref{dataset-stat}.
The entire dataset was then split into three subsets for training (90\%, 27,432), validation (5\%, 1,524) and testing (5\%, 1,525). 
We show two complete samples \footnote{\url{https://github.com/anonymous-tmp/anonymous-1}} of our dataset to expose how the judges hear the case and summarize factfinding in a practical situation. 

\begin{table}
\centering
\normalsize
\caption{Statistics of the dataset.}
\begin{tabular}{|l|l|}
\hline
avg. trial length (tokens)  & 723.5 \\
\hline
avg. factfinding length (utterances) & 38.6 \\
\hline
avg. factfinding length (tokens) & 158.8 \\
\hline
avg. expectant aspects & 5.4 \\
\hline
avg. missing entities & 2.2 \\
\hline
\end{tabular}
\label{dataset-stat}
\end{table}

\subsection{Implementation Details}
We use PyTorch 1.4 and AllenNLP 1.0 for implementing our models. 
For DIS, the vocabulary size is constrained to 20,000 for both source and target. 
For the trial text, we limit the utterance number to 50 and utterance length to 30. 
We train the word embeddings from scratch with embedding dimensions of 300; 
as reported by \cite{cohan2018a}, we also find no further improvement from pre-trained word embeddings. 
We set all the encoder hidden states to 256 dimensions. 
For the PEGASUS-DIS, we initialize the encoder-decoder with a PEGASUS model, which has 8 layers and 6 heads.

We use AdamW \cite{loshchilov2019decoupled} to optimize our models. 
For DIS and PEGASUS-DIS, the learning rate is set to 0.001 and 0.0003, respectively. 
For training, we set the batch size to 16 and 8 for the two models and group the instances into batches according to their padding lengths. 
For DIS, the hyperparameters in the final objective function are tuned as $\lambda_{a} = 0.5$, $\lambda_{m} = 0.35$. 
For PEGASUS-DIS, these two hyperparameters are set to 0.1 and 0.1, respectively.

We train our models on a single NVIDIA GeForce RTX 3090 GPU and use loss on the validation set to apply early stopping. 
It takes the two models about 9 epochs and 50 epochs for convergence. For inference, we use the beam search with a size of 5 to generate the summary. We set the maximum decode length to 200, which outnumbers 80 percent of all summary lengths in our dataset.

We are planning to release the dataset and code for further research on the problem. 

\begin{table*}[t]
\centering
\normalsize 
\caption{Results of summarization approaches on the judicial DIS dataset. The first and second section show extractive approaches' results and non-pretraining abstractive baselines' results. 
The third section shows the results of DIS in the non-pretraining setting and the last section compares our method with the SOTA baseline in the pretraining setting. 
}
{
\begin{tabular}{l|p{1.5cm}<{\centering}|p{1.5cm}<{\centering}|p{1.5cm}<{\centering}|p{1.5cm}<{\centering}|p{1.5cm}<{\centering}|p{1.5cm}<{\centering}}
\toprule[1.5pt]
\multirow{2}{*}{Methods} & \multicolumn{3}{c|}{ROUGE} & \multicolumn{3}{c}{BERTScore} \\
\cline{2-7} & 1 & 2 & L & $\rm P$ & $\rm R$ & $\rm F_1$  \\
\midrule[0.8pt]
LEAD3 & 15.39 & 3.19 & 10.13 & 68.25 & 66.92 & 67.49\\
TextRank & 20.91 & 6.69 & 13.27 & 71.91 & 70.70 & 71.21\\
\midrule[0.8pt]
S2S & 41.73 & 19.77 & 32.39 & 84.08 & 79.04 & 81.37\\
S2SAttn & 42.24 & 21.12 & 34.23 & 84.99 & 78.79 & 81.64 \\
PGN & 42.28 & 22.38 & 34.52 &  84.97 & 78.55 & 81.54 \\
FastSum & 45.91 & 23.37 & 32.72 & 83.84 & 79.70 & 81.58\\
DAH & 45.55 & 23.63 & 40.86 & 85.30 & 81.01 & 83.01 \\
\midrule[0.8pt]
DIS w/o EFAR & 47.56 & 27.28 & 38.09 & 85.19 & 81.21 & 83.04 \\
DIS w/o MFED & 48.53 & 28.09 & 39.24 & 85.25 & 81.67 & \textbf{83.32} \\
DIS & \textbf{49.02} & \textbf{28.67} & 39.78 & 84.93 & \textbf{81.98} & \textbf{83.32} \\
\midrule[0.8pt]
PEGASUS & 54.96 & 34.94 & 49.94 & 88.04 & 85.25 & 86.54 \\
PEGASUS-DIS & \textbf{55.64}  & \textbf{35.50} & \textbf{50.16} & \textbf{88.13} & \textbf{85.51} & \textbf{86.73} \\
\bottomrule[1.5pt]
\end{tabular}
}
\label{MainResults}
\end{table*}

\subsection{Baselines}
\label{baseline}
To demonstrate the effectiveness of DIS, we implement the following summarization models for comprehensive comparison:
\begin{itemize}[leftmargin=*]
\setlength{\itemsep}{0pt}
\setlength{\parsep}{0pt}
\setlength{\parskip}{0pt}
    \item \textbf{LEAD3} is a basic extractive baseline by selecting the first three utterances as the summary.
    
    \item \textbf{TextRank} \cite{mihalcea2004textrank} proposes an unsupervised extractive method based on sentence importance ranking. In the experiment, we extract the top-4 important sentences from trial.
    
    \item \textbf{S2S} \cite{sutskever2014sequence} is a basic encoder-decoder model for sequence-to-sequence learning. We concatenate all the utterances in the trial as the input sequence. 
    
    \item \textbf{S2SAttn} \cite{nallapati2016abstractive} modifies Seq2Seq model with the attentional decoder, which utilizes the context information given the attention distribution over encoder hidden states.
    
    \item \textbf{PGN} \cite{see2017get} extends attentional Seq2Seq model with copy mechanism. The decoder calculates a generation probability to choose between generating tokens from the fixed vocabulary and copying tokens from the source. 
    
    \item \textbf{FastSum} \cite{chen2018fast} jointly combines content selection and rewriting into a summarizer. We adjust the sentence-level content selection to utterance-level in the experiment.
    
    \item \textbf{DAH} \cite{cohan2018a} proposes a model driven by hierarchical attention. It is suitable for the discourse structure of judicial trials.

    \item \textbf{PEGASUS} \cite{DBLP:conf/icml/ZhangZSL20} is a Transformer-based abstractive summarization model which is pre-trained on large text corpora with a self-supervised objective, which removes/masks important sentences from the original documents and then generates them. To make a fair comparison, we fine-tune the pre-trained PEGASUS on the PLD dataset with exactly the same hyperparameters we used to train PEGASUS-DIS.
\end{itemize}

\begin{table}
\centering
\normalsize
\caption{Human evaluation results.}
{
\begin{tabular}{l|c|c|c}
\toprule[1.5pt]
Summary & L.C. & F.C. & Read.\\
\midrule[0.8pt]
S2SAttn & 2.32 & 1.22 & 4.22 \\
DAH & 3.18 & 2.65 & 4.45 \\
DIS & 3.60 & 3.33 & 4.50 \\
PEGASUS & 3.91 & 3.84 & 4.68 \\
PEGASUS-DIS & {\bf 3.99} & {\bf 3.93} & {\bf 4.72} \\
\midrule[0.8pt]
Ground Truth & 4.70 & 4.25 & 4.88 \\
\bottomrule[1.5pt]
\end{tabular}
}
\label{HumanEval}
\end{table}

\subsection{Evaluation Metrics}
\label{metrics}
To validate the overall quality of the generated factfinding, we report results on three evaluation metrics for different perspectives: (1) ROUGE \cite{lin2003automatic}, (2) BERTScore \cite{zhang2020bertscore} and (3) human evaluation.

ROUGE is the standard metric for summarization task. We report the $F_1$ score for ROUGE-1, ROUGE-2, and ROUGE-L, obtaining the scores with the files2rouge\footnote{\url{https://github.com/pltrdy/files2rouge}} package. From a linguistic perspective, the indicators can reflect the informativeness and fluency of the summary.

BERTScore is recently proposed as a novel evaluation metric for text generation tasks. Instead of the exact token match in ROUGE, BERTScore computes token similarity with contextualized embeddings from pre-trained language models like BERT \cite{devlin2019bert}. In the experiment, we choose 12-layer $\rm BERT_{BASE}$ to get the score.

We perform human evaluation to verify whether the performance on automatic metrics correlates with human perceived quality. We hire five annotators well trained in reading legal documents and randomly select 100 cases from the test set for evaluation. For each case, we show the judicial trial, the factfinding in the verdict, and results generated by S2SAttn, DAH, and DIS. The annotators are asked to score each factfinding on three indicators under the 1 to 5 scale: Logical Completeness, Factual Consistency, and Readability.
\section{Result Discussion}
\subsection{Overall Performance}
\begin{figure*}[t]
\centering
\includegraphics[width=\linewidth]{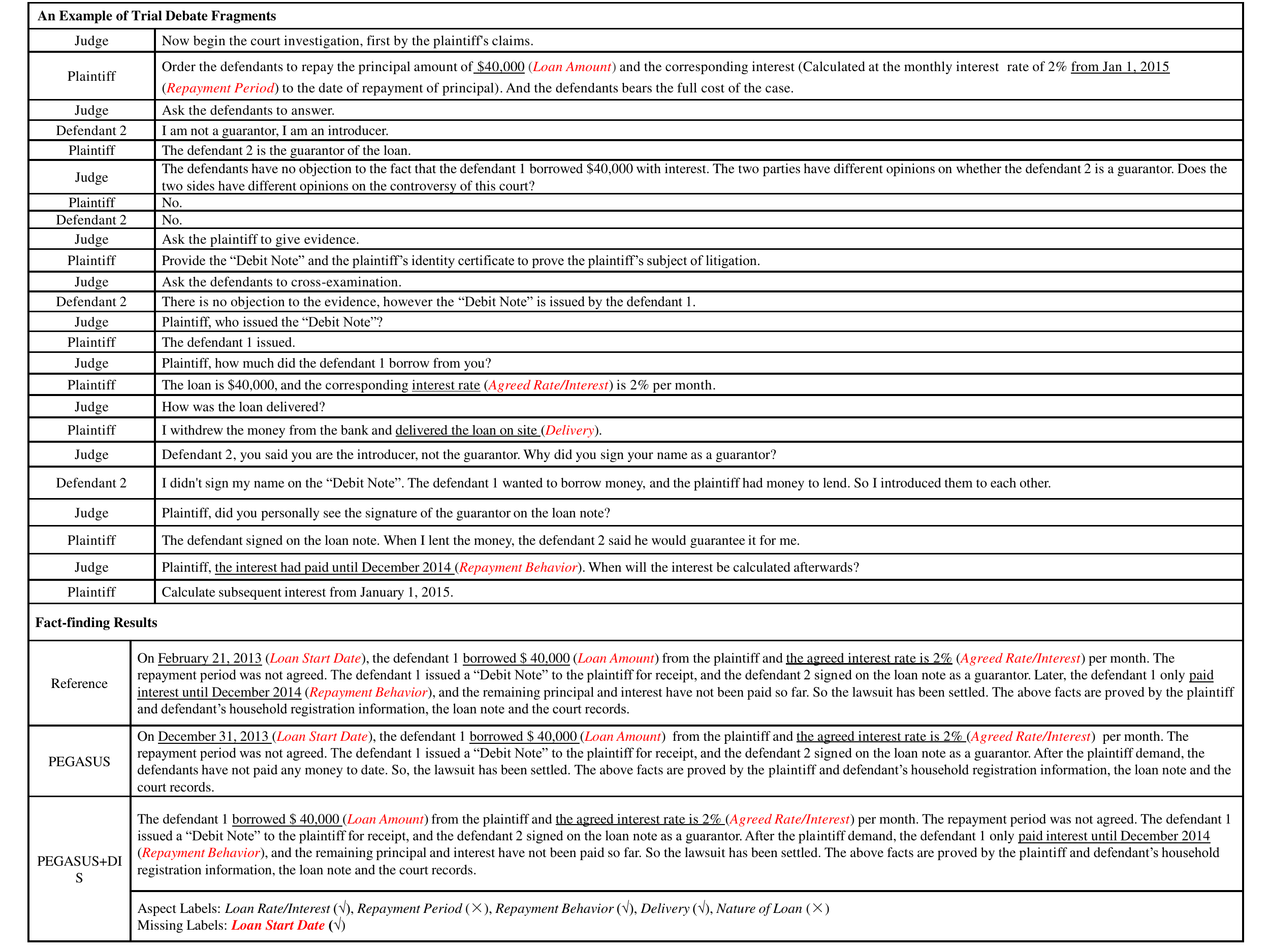}
\caption{Case Study 1. Words in red denote the factual aspects that are correctly generated by the tested methods while the words in green denote the incorrect ones. Our proposed model also generate inspectional summarization with the predicted essential aspects to be contained in the summary as well the missing entity (\textit{Loan Start Date} in orange). The \checkmark and $\times$ indicate whether the generated aspects and missing entities are evaluated correctly or not.}
\label{caseStudy}
\end{figure*}

We evaluate the overall performance comprehensively from the following perspectives: (1) comparison against baselines on automatic metrics, 
(2) the effectiveness of proposed auxiliary tasks, and (3) human perceived quality of generated summaries.

\textbf{Comparisons against baselines.} As shown in Table~\ref{MainResults}, we first report the experimental results against baselines on automatic metrics. 
Following the setting in section~\ref{baseline} and \ref{metrics}, we have the observations based on the results: (1) Our DIS model outperforms all the baselines without pre-training over ROUGE-1, ROUGE-2 and BERTScore, though slightly worse than DAH over ROUGE-L. 
(2) The fine-tuned PEGASUS model outperforms all the models without pre-training, which shows the advantages of large pre-trained models. 
(3) However, the PEGASUS-DIS model still gives better results than PEGASUS on all automatic evaluation metrics, which demonstrates that logical/factual gaps may still exist in pre-trained models and the DIS framework is still usefull in this setting. 
(4) The two basic extractive methods, LEAD3 and TextRank, perform rather poorly on the PLD dataset because the factual details scattered in the dialogue are difficult to represent by certain utterances. DAH performs better than other abstractive baselines, indicating that the hierarchical representative learning is critical in summarizing dialogues. 

\textbf{Human perceived quality.} As shown in table~\ref{HumanEval}, we report the human evaluation results. 
First, DIS outperforms the selected baselines without pre-training on all the indicators, especially in factual consistency, while still having a certain gap with the reference summary. It is noteworthy that S2SAttn performs poorly in factually related indicators despite having acceptable performance in terms of automatic evaluation. This finding indicates that existing evaluation metrics are insufficient to tell the differences concerning factuality. 
Second, The fine-tuned PEGASUS  model  again demonstrates its advantages in human evaluation experiments on the three indicators, which demonstrate that pre-trained models can not only improve the fluencies but also be able to relieve the logical/factual inconsistency to some extent. Third, the proposed PEGASUS-DIS still outperforms PEGASUS on the three indicators, which again shows that the DIS framework is able to benefit pre-trained abstractive summarization models.
Finally, as for readability, the level of all the results is relatively high, without obvious distinction. It reflects that generating factual consistent summaries is more significant than generating fluent summaries.

\begin{figure*}[t]
\centering
\includegraphics[width=\linewidth]{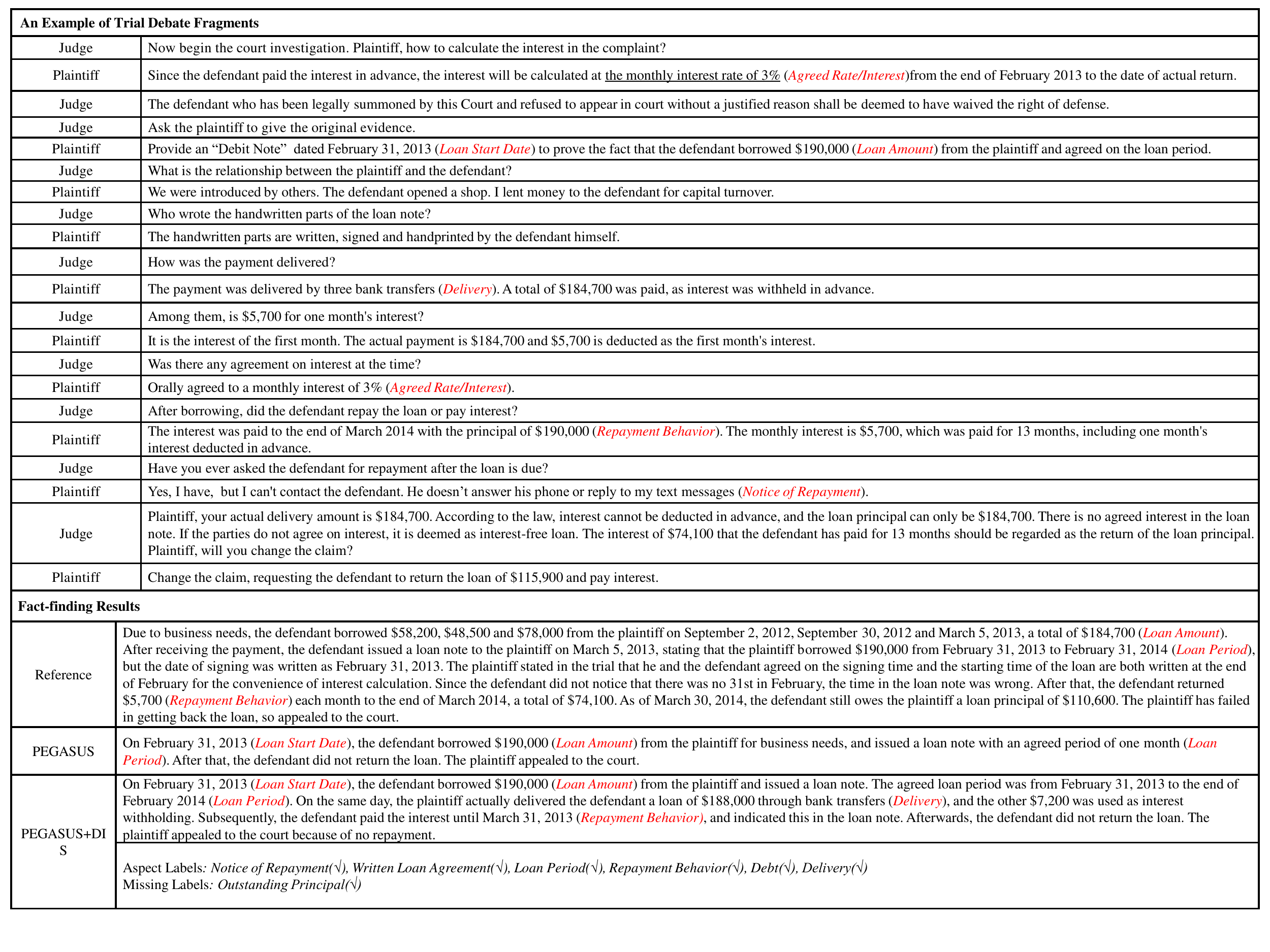}
\caption{Case Study 2. Words in red denote the factual aspects that are correctly generated by the tested methods while the words in green denote the incorrect ones. Our proposed model also generate inspectional summarization with the predicted essential aspects to be contained in the summary as well the missing entity (\textit{Loan Start Date} in orange). The \checkmark and $\times$ indicate whether the generated aspects and missing entities are evaluated correctly or not.}
\label{caseStudy2}
\end{figure*}

\textbf{Effectiveness of auxiliary tasks.} We conduct the ablation test to assess the contribution of each component in DIS framework. The fourth section in table~\ref{MainResults} shows the results. We maintain the aspect embedding to compute the aspect-aware decoder state when removing EFAR task. Its removal causes 4.8\% relative increase in error (RIE) for ROUGE-2 $F_1$ score, of which the impact is more critical than the removal of MFED task (2.0\% RIE for ROUGE-2 $F_1$ score). The finding indicates that both the auxiliary tasks contribute to the informativeness, while EFAR presents a more significant role.

\subsection{Case Study}

Figure~\ref{caseStudy} and ~\ref{caseStudy2} show two case studies by comparing the generated summary of the proposed method against the best baseline and the reference summary. 
As shown in Figure~\ref{caseStudy}, our method outperforms PEGASUS in generating high coverage of accurate factual content by predicting the expectant factual aspects correctly. For example, the proposed PEGASUS-DIS predicts the Repayment Behavior correctly and generates corresponding texts in the summary. Also, there exists one missing factual entity in the reference summary (Loan Start Date). DIS detects the inconsistency to inform the users.
Similar observations can be found in Figure\ref{caseStudy2}. With the help of the EFAR auxiliary task, the PEGASUS-DIS generates factual contents, e.g., Repayment Behavior and Delivery, which are missed by the fine-tuned PEGASUS.

\subsection{Error Analysis}
From the sampling statistics and human evaluation feedback, we conclude the major problems occurring in DIS results as follows: (1) 25.5\% of the errors come from the failure to distinguish between multiple facts. The lengthy and colloquial dialogue makes it difficult for the model to capture subtle differences and determine the correct facts.
(2) 18.1\% of the errors are attributed to their adverse results on auxiliary tasks. EFAR and MFED predictions are related to the summary content so that the false aspect can mislead to false facts.
(3) We also find that 14.6\% of the errors happen when the factual information needs to be obtained through a certain degree of reasoning. For instance, the summary may infer a specific date or calculate the total amount given fragmented information in dialogue. To improve the proposed method in the future, introducing numerical calculation and reasoning capabilities into text generation can be a promising direction.

\section{Related Work}
Generally, neural text summarization has two strategies: extractive summarization and abstractive summarization. Extractive summarization merges the selected utterances from dialogue to form the summary, so it fails to produce consistent results. For dialogue inspectional summarization, abstractive strategy is more suitable because it can generate novel expressions to summarize the interactions.
In the past few years, neural abstractive text summarization has received much attention from the research community. Seq2Seq \cite{sutskever2014sequence} framework enables the summarizer to generate diverse, relevant, and readable results. Since the first successful attempt by \cite{rush2015a}, many techniques have been proposed to improve the accuracy and versatility. 
The most influential work includes copying mechanism \cite{gu2016incorporating, see2017get},  AMR parsing \cite{liu-etal-2015-toward}, reinforcement learning approaches \cite{paulus2018a,chen2018fast}, coverage \cite{tu-etal-2016-modeling} and combination of extraction and abstraction \cite{gehrmann2018bottom, hsu-etal-2018-unified, celikyilmaz-etal-2018-deep}.

Our work especially belongs to dialogue summarization, which has presented new challenges to abstractive summarization: information in the interactions are more difficult to capture than documents with sequential logic; the lack of suitable corpora restricts the relevant research progress. 
Recent work adapted the practice in news summarization to this field and introduced supplementary annotations. 
For instance, \cite{goo2018abstractive} incorporates dialogue acts to better recognize the interactive patterns in meeting summarization; 
\cite{liu2019automatic} introduces key point sequence generation to improve the logic and integrity of summaries in the customer service domain.
\cite{DBLP:conf/cikm/DuanZYZLWWZS019} pay attentions to the controversy focus space of a civil trial dialogue. 
Unlike their work, our proposed framework particularly focuses on the factual inconsistency problem, aimed at generating a more faithful summary.

More recently, pre-trained models based on Transformer \cite{DBLP:conf/nips/00040WWLWGZH19, DBLP:conf/icml/SongTQLL19, liu-lapata-2019-text} have shown their dominant advantages from language understanding tasks to language generation tasks, e.g., machine translation, dialogue generation and text summarization. 
For language generation pre-training, a core problem is how to construct the self-supervised task from the unlabeled corpus.
UniLm \cite{DBLP:conf/nips/00040WWLWGZH19} proposed a unified model for both language understanding and generating tasks by designing different self-attention masks for different kinds of tasks.
T5 \cite{JMLR:v21:20-074} converts various NLP tasks into a unified text-to-text framework, and trains the tasks with a sequence-to-sequence model.
MASS \cite{DBLP:conf/icml/SongTQLL19} is a sequence-to-sequence pre-training model, which masks a consecutive text fragment in a sentence and predicts the masked tokens using a encoder-decoder model.
BART \cite{lewis-etal-2020-bart} constructs the self-supervised task by corrupting the document with transformations, e.g., token deletion and text infilling, and then recovering the original document.
PEGASUS \cite{DBLP:conf/icml/ZhangZSL20} is the state-of-the-art abstractive summarization model, which is pre-trained on a large-scale corpora by generating the selected important sentences as a self-supervised task.
Our work further investigates if such pre-trained abstractive summarization models have the factual inconsistency problem, and if the proposed DIS framework can benefit the pre-trained models.


\section{Conclusion}
In this work, we mainly investigate Dialogue Inspectional Summarization (DIS) by solving factual inconsistency problem. 
We propose DIS as a novel end-to-end dialogue summarization framework under non-pretraining and pretraining settings, supervised with two auxiliary tasks, namely Expectant Factual Aspect Regularization (EFAR) and Missing Factual Entity Discrimination (MFED). The auxiliary tasks align the generated summary with dialogue on different factual granularity.

For experiments, we benchmark the DIS dataset in judical trial summarization. Based on comprehensive evaluation and analysis, we demonstrate that the DIS framework can generate a more readable summary with accurate coverage of factual aspects, as well as informing the user with potential factual inconsistencies for further human intervention.

\bibliography{IEEEexample}
\bibliographystyle{IEEEtran}

\end{document}